# FARSPREDICT: A BENCHMARK DATASET FOR LINK PREDICTION

**Najmeh Torabian**
Department of Computer Engineering
Central Tehran Branch, Islamic Azad University
Tehran, Iran
najmeh.torabian@gmail.com

**Behrouz Minaei-Bidgoli**
Department of Computer Engineering
Iran University of Science and Technology
Tehran, Iran
b_minaei@iust.ac.ir

**Mohsen Jahanshahi**
Department of Computer Engineering
Central Tehran Branch, Islamic Azad University
Tehran, Iran
mjahanshahi@iauctb.ac.ir

## ABSTRACT

Link prediction with knowledge graph embedding (KGE) is a popular method for knowledge graph completion. Furthermore, training KGEs on non-English knowledge graphs promote knowledge extraction and knowledge graph reasoning in the context of these languages. However, many challenges in non-English KGEs pose to learning a low-dimensional representation of a knowledge graph's entities and relations. This paper proposes "Farspredict," a Persian knowledge graph based on Farsbase (the most comprehensive knowledge graph in Persian). It also explains how the knowledge graph structure affects link prediction accuracy in KGE. To evaluate Farspredict, we implemented the popular models of KGE on it and compared the results with Freebase. Given the analysis results, some optimizations on the knowledge graph are carried out to improve its functionality in the KGE. As a result, a new Persian knowledge graph is achieved. Implementation results in the KGE models on Farspredict outperforming Freebase in many cases. At last, we discuss what improvements could be effective in enhancing the quality of Farspredict and how much it improves.

## 1 Introduction

Knowledge graphs have received much attention in recent years due to their applications that offer significant economic benefits. A Knowledge graph contains the knowledge obtained from the sources, including texts and tables. It has many applications in natural language processing and has been investigated as a potential reasoning source for explainable artificial intelligence.

Although the impact of creating knowledge graphs in non-English languages has been explored recently, little attention has been paid to preparing a suitable knowledge graph for use in the link prediction field.

At the same time, one of the main reasons that significant progress has yet to be made in Persian reasoning, recommendation systems, and other similar fields is the need for a proper knowledge graph in these languages. Although some attempts have been made to construct a Persian knowledge graph, the most successful is the Farsbase project. By applying Farsbase for link prediction through KGE models, we realized it is too weak to be used for link prediction.

In approach to state-of-the-art link prediction methods, we come to the KGE methods. These methods were introduced with TransE, which falls into translational distance models. After TransE [Bordes et al., 2013], the TransH [Wang et al., 2014], TransD [Ji et al., 2015], TransR [Lin et al., 2015] models, and many other methods improved TransE and its other models. These methods convert triples into vectors and predict the new triple in the Freebase and WordNet graphs by considering the location of the embedded vectors. Almost another group of models known as semantic similarity emerged simultaneously with the translational distance models, including RESCAL [Nickel et al., 2011], DistMult [Yang et al., 2015], HolE [Nickel et al., 2016], and ComplEx [Trouillon et al., 2016]. Similar to the previous category, these models mostly use Freebase, and WordNet datasets apply matrix decomposition techniques to embed existing triple components and predict new triples. The latest and most widely used category of link prediction models in KGE methods is deep learning methods.

A Persian dataset is rarely used experimentally for link prediction despite its acknowledged importance. Therefore, considering the practical application of these link prediction methods, we decided to survey the implementation of these models on non-English and especially Persian knowledge graphs. Unfortunately, few were found in low-resource languages, but no case was found in Persian.

This study aims to develop a Persian knowledge graph for link prediction and use it through the KGE models for knowledge graph completion as a valuable knowledge reference and obtain embedded vectors to be applied in Persian applications. Therefore, the current paper presents a routine in some steps to make changes in Farsbase and make a standard knowledge graph for link prediction. Based on this routine, it then describes the implementation of the KGE models on the prepared knowledge graph to find out how reliable this dataset could be in this field.

The paper is structured as follows. Section 2 deals with the related works, and Section 3 deals with the proposed three-step standardizing of our Persian knowledge graph and implementing the KGE models. Section 4 discusses experiments on Farspredict and the results incorporated in this section. Quality analysis is composed in Section 5. Finally, the paper concludes the work in Section 6, along with the future research scope.

## 2   Related Works

Since knowledge graphs are created to understand and reason human languages, human knowledge needs to be represented, stored, and extracted in a form that computers can process. The knowledge graphs have been developed as a knowledge base for entities and relations among these entities. However, they have a limitation in coping with most human languages.

On the other hand, KGE has provided extraordinary progress in representation over the years, but many still need to apply it to real datasets. The size, intricacy, and variety significantly challenge their utilization.

Several partly human efforts have been invested in making KGs available across languages [Lakshen et al., 2018] [Fafalios et al., 2023]. However, even well-known KGs, including DBpedia [Auer et al., 2007], Wikidata [Vrandečić and Krötzsch, 2014], YAGO [Tanon et al., 2020], Freebase [Bollacker et al., 2008], BabelNet [Navigli and Ponzetto, 2012], and Google knowledge graph [Ehrlinger and Wöß, 2016], are most abundant in their English version. This lack of a multilingual knowledge graph limits the porting of natural language processing (NLP) tasks, such as link prediction, question answering, and recommendation systems, to different languages. However, several attempts have been made to construct knowledge graphs in other languages; examples are given below. The remarkable thing about these knowledge graphs is that there is no report indicating the use of these items in KGE and link prediction.

XLore [Wang et al., 2013] is an English-Chinese bilingual knowledge graph that enriches Chinese knowledge using online wikis. XLore's authors tried covering semantic inconsistency between concepts, not equivalent cross-lingual entity linking problems. XLore2 [Jin et al., 2019] is an XLore extension established to eliminate the incompleteness of the dataset by adding facts via making cross-lingual knowledge linking, cross-lingual property matching, and fine-grained type inference. The following non-English knowledge graph is Lynx [Montiel-Ponsoda and Rodríguez-Doncel, 2018]. The legal knowledge graph in the multilingual European knowledge graph is part of a big project for innovative compliance services. Lynx project benefits from the technology-driven content product of KDictionaries, which developed quality lexicographic data in 50 languages. Besides, HOLINET [Prost, 2022] is a Holistic KG for French, PolarisX [Yoo and Jeong, 2020] is an automating expansion KG, RezoJDM [Mirzapour et al., 2022] in French, and VisualSem [Alberts et al., 2021] is a visual multilingual knowledge graph.

Due to the increasing applications of knowledge in various aspects of human life, downstream requests are increasing. The number of specific knowledge graphs is also growing. To clarify this article, we will explain some examples of these applications below.

**The cultural heritage** Chinese knowledge graph [Fan and Wang, 2022], which can grow, provides updated information to its users. There is also another article in the Chinese language [Liu et al., 2020] that collects information automatically with a crawler and uses machine learning tools to build this graph. Research [Marchand et al., 2020] also includes information on the cultural heritage of the province of Quebec in French. In addition to the mentioned cases, the article of Ms. Bruns and her colleagues [Bruns et al., 2021] to preserve the culture of Nuremberg can be mentioned. Also, research [Tan et al., 2021] has been presented to protect the spread of the book's heritage in German. Arco [Carriero et al., 2019] is another European knowledge graph in the Italian language. Indeed, the Arco knowledge graph is a product of a Cultural Heritage project, including software, a documentation report, and some more parts.

**Medicine** A particular purpose knowledge graph called TCMKG [Zheng et al., 2020] was created by zheng and his colleagues to preserve traditional Chinese medicine. The situation of the emic of COVID-19 inspired the construction of a knowledge graph based on the spatial distribution of this disease in the article [Yang et al., 2022]. In addition to this article, other research has been done in [Feng et al., 2022] and [Sakor et al., 2023] that have built knowledge graphs based on the distribution of the COVID-19 vaccine and disease. The content of this dataset is obtained by primarily establishing countries with this vaccine's research and development team extracting entities and relations.

**Other applications** In addition to the above, other datasets exist in other applications. Such as WeaKG-MF [Ayadi et al., 2022], a knowledge graph containing collected meteorological data in French, a particular knowledge graph for the application of question answering in solving problems of electrical devices in Chinese [Meng et al., 2022], a knowledge graph containing information from various data sources scattered in European Union [Soylu et al., 2020] research [Mishra et al., 2022] is a specific knowledge graph for the tourist industry. And finally, the article [Zehra et al., 2021] provides an automatic query engine to find hidden relationships between financial documents.

In addition to the mentioned knowledge graphs, CAMS_KG [Bounhas et al., 2020] is a morpho-semantic knowledge graph that combines Ghwanmeh stemmer and MADAMIRA to support Arabic knowledge representation and information retrieval. Since Arabic is one of the languages closest to Persian, this study is essential. Besides, the last and most crucial non-English knowledge graph for us is Farsbase [Asgari-Bidhendi et al., 2019]. Farsbase is the first Persian knowledge graph for extracting knowledge from multiple sources. Farsbase obtains a rich knowledge graph from popular sources, including Wikipedia and online sites. This excellent knowledge graph is the primary source of the present research, although it has some problems to be used in the link prediction field.

Although many projects have been done to create non-English or multi-language knowledge graphs, no reliable Persian knowledge graph was found for link prediction. To the best of our knowledge, the literature has yet to discuss a Persian knowledge graph using knowledge graph representation and link prediction using the KGE models.

Despite modern techniques for constructing knowledge graphs from primary sources, such as text and tables, in most cases, knowledge graphs will have a partial result. So to extract knowledge from them, they need to be completed through methods, including link prediction.

The KGE models are a way to embed a knowledge graph in a vector space while retaining its main properties. These models are divided into three main categories, some of which are briefly discussed below. In these models, a triple is proposed to be added to the knowledge graph to complete the dataset through link prediction techniques, such as negative sampling [Bharambe et al., 2023].

**Translational distance models** are the first group of embedding models, in which each of the three elements (head, relation, and tail) is considered as a vector in the embedded space according to the chosen specific model, and the space of the elements can be shared or separate [Wang et al., 2017]. The first model in this category is the TransE [Bordes et al., 2013], which has achieved remarkable success. The scoring function used in this model is $h + r \approx t$. The main disadvantage of this model is that it covers only 1-to-1 triples. Other models include TransH [Wang et al., 2014], TransR [Lin et al., 2015], TransD [Ji et al., 2015], and several other models that try to solve this problem and improve the efficiency of TransE results and its other models.

**Semantic matching models** are based on semantic similarity matching and the relations between entities through the scoring function. Scoring functions in this model are essential in identifying new triples and introducing them to the knowledge graph. The first model in this set is RESCAL [Nickel et al., 2011], in which entities are considered vectors, and the relations between these entities are matrix. Other models in this category try to improve the performance of the RESCAL model, such as DistMult [Yang et al., 2015], HolE [Nickel et al., 2016], and ComplEx [Trouillon et al., 2016]. In these models, relational data is modeled as a three-way tensor X of size n×n×m, where n is the number of entities and m is the number of relations. This knowledge representation model limits the scope of knowledge graph completion within the knowledge graph and cannot predict new knowledge beyond the knowledge graph.

**Neural network models** use deep neural network tools to learn the ternary model in a knowledge graph. Then, through the exploratory structure, propose new triples and complete the knowledge graph. One of the first methods to embed the knowledge graph based on deep learning is the ProjE model [Shi and Weninger, 2017], in which entities and relations between them are embedded seamlessly and cohesively. This model emphasizes reducing the number of parameters, leading to less time complexity and less computation. Other models in this category include ConvKB [Nguyen et al., 2019], ConvE [Dettmers et al., 2018], RotatE [Sun et al., 2019], SACN [Shang et al., 2019], and many more. The deep neural network models have shown good predictive performance, even if they are more expensive and time-consuming compared two other categories.

In this research, Farsbase is considered a primary dataset on which the KGE models are implemented. The results of an empirical study showed that Farsbase has some inconsistencies to be used in link prediction and brought the idea of making a new knowledge graph for KGE models-based link prediction.

## 3 Persian knowledge graph construction

Knowledge graphs are valuable resources that, in the first place, provide the possibility of extracting knowledge from textual sources. In the second place, they are used as a data reference for many technologies, including question-answering systems, recommender systems, and knowledge management. Therefore, this type of data reference is necessary for every language. Farsbase, as the first Persian graph knowledge, is a rich source, but it could be more effective for use in link prediction and completing graph knowledge, and its shortcomings cause knowledge extraction to be disturbed. The process used to create the new knowledge graph is discussed in this section.

Two hypotheses are proposed with the studies on the structure of graph knowledge and knowledge graph embedding in the form of multiple models. The first hypothesis: by creating a regular knowledge graph in the Persian language, it is possible to achieve link prediction results similar to the outputs of KGE models. The second hypothesis: Since the reports of deep learning methods have good performance among knowledge graph embeddings, by implementing deep learning models on the obtained Persian knowledge graph, better accuracy was achieved than other methods in link prediction.

In order to implement KGE models on the Persian dataset and evaluate the dataset performance, we used the OpenKE framework, which contains some graph knowledge embedding models [Han et al., 2018]. The source code of this framework is open, and at the time of writing this article, nine valid KGE models have been developed and are available in this framework. In this framework, two datasets Freebase [Bollacker et al., 2008], and WordNet [Fellbaum, 2010], have been used in research and implementation. We implemented the models in this framework on Farsbase graph knowledge. The results were significantly weaker than the results reported in the articles related to the embedding models of the knowledge graph. In order to fix the shortcomings, we created a new Persian knowledge graph for link prediction based on the Farsbase dataset. In the following, we will describe the specifications of Farsbase and introduce Farspredict.

### 3.1 Farsbase Specifications

We need to implement the KGE models on our Persian knowledge graph to obtain embedded vectors of the knowledge graph for link prediction. Since Farsbase is the first Persian knowledge graph, this dataset was used to get the embedded knowledge graph. However, preparing a version of the knowledge graph was necessary to do this. The cause of the first changes refers to the Farsbase ontological structure. We need triples with two entities and a relation between them, not properties or anything else. The specifications of this version are listed in Table 1.

Table 1: Specifications of a version of the Fars- base Knowledge Graph with two entities

| Dataset | # relations | # entities | # triples |
|---|---|---|---|
| **Farsbase** | 7378 | 541927 | 2398999 |

Although Farsbase is a rich Persian knowledge graph prepared from unchecked and tabled unsupervised texts, many of its triples do not represent a fact and have no inferential value. However, a large number of relations in Farsbase have few facts. Actually, 71% (1392/392) of relations have facts less than 100. Many of them are limited to having only one fact per relation type. That is the cause of a large and sparse knowledge graph.

The next step for providing a representation of the Persian knowledge graph and comparing the results with standard datasets is implementing OpenKE models on the proposed version of Farsbase. According to the structure required to implement the KGE models, we divided this version of the dataset extracted from Farsbase into three training, testing, and validation sections and then applied the KGE models to them. In this phase, 10% of the dataset was allocated for the validation dataset, 20% for the testing dataset, and 70% for the training dataset. We applied six translational distance models, including TransE, TransH, TransR, TransD, TransG [Xiao et al., 2016], and TransM [Fan et al., 2014], and also four models of semantic matching models, namely: RESCAL, HolE, ComplEx and DistMult.four The output is based on two metrics, Hits@10 and Mean Rank achieved. Indeed, the mean rank is the average rank for all predicted triples within each model $((1 + n))/2$, and the proportion of testing triples whose ranks are not larger than 10 is HITS@10. That is called the "Raw" setting. It is the "Filter" setting when we filter out the corrupted triples in the training, validation, and testing datasets. If a corrupted triple exists in the knowledge graph, ranking it before the original triple is acceptable. Higher HITS@10 and a lower mean rank mean excellent performance.

The results from the implementation of models on Farsbase are too weak compared to the same models on the Freebase and WordNet datasets. The degree of nodes in the Farsbase graph is very different in that many entities and relation types are seen in only one of the triples, while a small number of entities have a degree above 30,000. As explained in the "Related Works" (Section 2), the Farsbase dataset was prepared in RDF format and was very rich as the first Persian knowledge. However, it is not ready to implement link prediction applications and can only contain data as a reliable source.

### 3.2 Farspredict

In order to validate Farsbase in link prediction, we first had to make changes in this dataset to prepare it to use with KGE models. The routine described here is simple, rapid, and reliable in designing a knowledge graph called Farspredict. Therefore, a process consisting of the following steps was considered to create a standardized knowledge graph.

These modifications are based on KGE models to use knowledge graphs in link prediction. The removal of images or hyperlinks is for this purpose, indicating that the existence of the image and the hyperlink does not violate the properties of the knowledge graph.

**Steps of Farspredict standardization:** It includes three parts:

**Part one** Content modification of the triples.

Some entities are in English or Arabic, and some contain non-text items. In order to prepare content in this section, two steps of corrections are presented. Step 1: Entities in collected triples, images, or URLs. These triplets are removed. Second step: removing non-Persian entities and relation types. For example, "isA" and "isRelatedTo" in the knowledge graph are correct in terms of being triple in the knowledge graph but are omitted because they are non-Persian. The same procedure is performed for entities in the non-Persian language. After the implementation of this section, the number of triples decreased to 1.5 million.

**Part Two** Graph structural modification The existing graph is disconnected and has single nodes up to this point. We go through the following two steps to optimize the structure. Step 1: Evaluate the graph to find the number of unrelated subgraphs. At the end of this evaluation, it was found that this graph consists of 48 components, the largest subgraph of which accounts for 99.87% of all triples in the graph. Second step: we remove small subgraphs and single nodes.

**Part three** Content modification in the knowledge graph. In addition to being irregular, the graph obtained from the previous sections shows that the number of triples of different relation types is a vast range. To regularize the graph and better triples distribution in the graph, we present five optimization levels in this section; the order of execution is significant. Level one: We calculate the number of triples of each relation type. Level two: We removed less than 100 triples for each relation type. Level three: We removed the triples containing the deleted relations. Level four: removing isolated nodes (incoherent components) due to removing relations. Level five: We remove the entities and relations that no longer exist during graph standardization and triple elimination from the entities and relations repository. These steps are shown in Algorithm 1 and Figure 1.

Following the standardization steps, the first version of the new dataset, Farspredict, was created.

Implementation results were still far from expectation and weaker than Freebase results. Therefore, by examining the graph, we realized there is a chain in some parts of the current graph. To reduce graph chains, we removed entities with less than five connections (entities connected to less than five other entities) from the dataset. Finally, the specifications of the second version of the Farspredict knowledge graph were determined, as shown in Table 2.

Table 2: Specifications of the final version of the Farspredict

| Dataset | # relations | # entities | # triples |
| --- | --- | --- | --- |
| Farspredict | 392 | 107827 | 622287 |

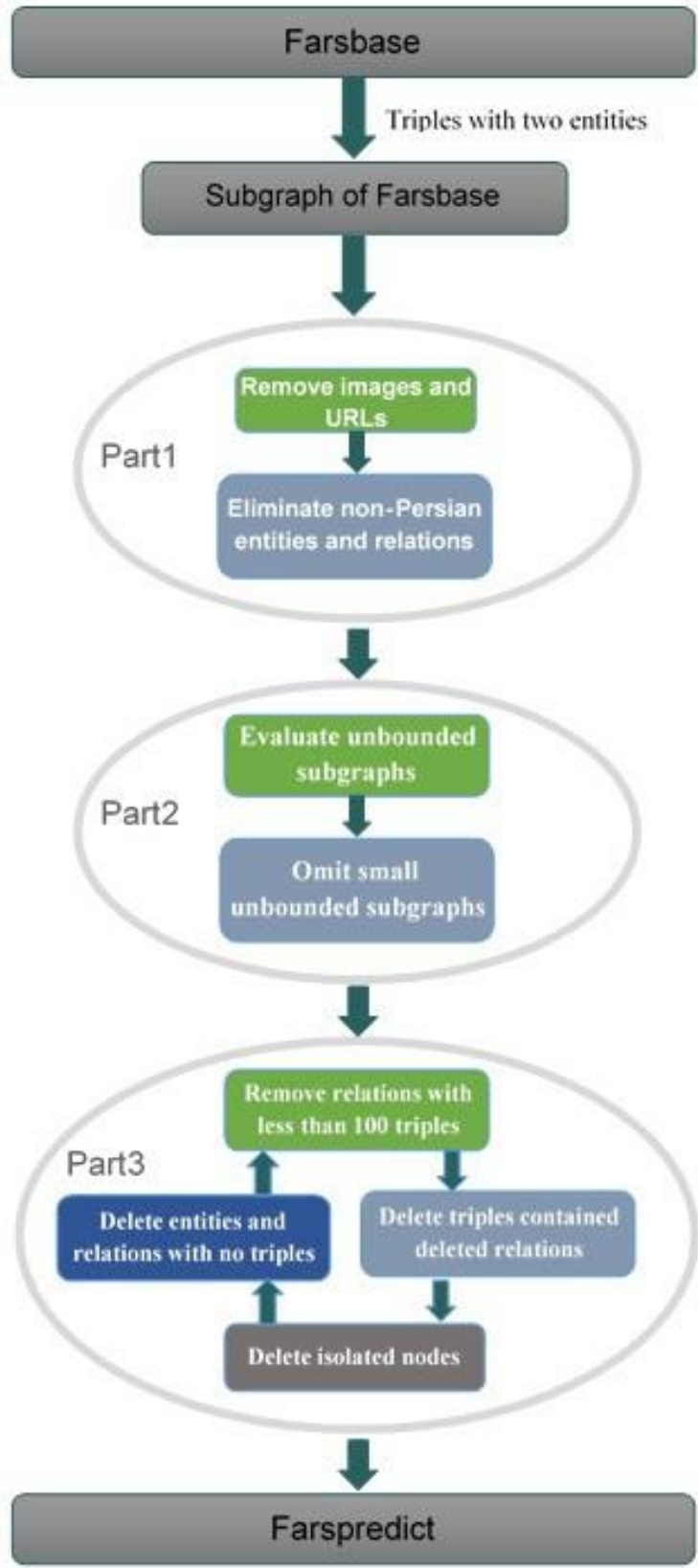

Figure 1: Standardization Schema

---

**Algorithm 1** Knowledge graph standardization
---
**Require:** Knowledge graph Farsbase,dts, Farspredict
**Require:** head h, tail t, relation r, triple tp, ent[], rel[]
1: **while** tp in Farsbase **do**
2:     **if** t and h in ent [] **then**
3:         $dts = dts \cup tp$
4:     **end if**
5: **end while**
6: Subgraphs = cgSpan [Shaul and Naaz, 2021](dts)
7: set FarsPredict = biggest subgraph in Subgraphs
8: Remove images and U RLs from ent[]
9: Remove non Persian r from rel[] and t, h from ent[]
10: Repeat lines 6 and 7
11: **while** r in rel[] **do**
12:     **if** $tp(r) < 100$ **then**
13:         Remove r from rel[]
14:         Remove tp(r)
15:     **end if**
16: **end while**
17: **while** e in ent[] **do**
18:     **if** factchecker[Shiralkar et al., 2017] (e) < 5 **then**
19:         Remove e from ent[]
20:         Remove tp(e)
21:     **end if**
22: **end while**
23: Repeat lines 6 and 7
24: **Return** Farspredict

---

## 4 Experiments and Results

In this section, we describe the experiments conducted on KGE models, using a Persian knowledge graph to provide embedded vectors of triples. It is expected that after going through the steps of constructing the Persian graph knowledge in the previous section, you will make significant progress and provide the prerequisites for link prediction.

We compare the knowledge graph created in Section 3 with the benchmark datasets. The ratio of the number of entities to the number of relation types in Freebase and Farspredict shows that the new graph is sparse and there are relation types with low frequency.

Therefore, the implementation results were far from expectations and weaker than the results of Freebase. We examined the graph and found a chain in some parts of the current graph. Consequently, we removed entities with fewer than five degrees (entities connected to fewer than five other entities) from the dataset to reduce graph chains.

**Human evaluation** First, we randomly select 1000 triples from Farspredict. We remove the tails of these triples. These incomplete triples are given to three human experts who have basic knowledge of knowledge graphs. The results obtained are 95.1, 93.4, and 93% accurate. The almost high average of 93.83% demonstrates that the dataset triples are mostly clear for humans.

**Application by link prediction** The KGE models must be implemented on this dataset to examine the new knowledge graph. We implemented the same procedure on the Freebase dataset using the OpenKE framework. The results are presented in Table 3.

As shown in Table 3, the results improved sufficiently, and the dataset results became closer to the standard dataset in the ratio of the number of relations to the number of triples. The Persian knowledge graph is still sparse despite general satisfaction with the results. We will use link prediction to eliminate the sparsity of the knowledge graph and complete it.

Table 3: Execution of KGE models on the final version of Farspredict

|         | Mean Rank |          | Top10  |          |
|---------|-----------|----------|--------|----------|
| Models  | Raw       | Filtered | Raw    | Filtered |
| TransE  | 3532.356  | 2980.389 | 0.318  | 0.374    |
| TransH  | 4509.083  | 3958.75  | 0.291  | 0.338    |
| TransD  | 4290.836  | 3732.386 | 0.294  | 0.343    |
| DistMult| 8176.77   | 7676.532 | 0.165  | 0.171    |
| ComplEx | 8228.179  | 7703.388 | 0.112  | 0.125    |
| RESCAL  | 96559.422 | 96439.093| 0.0004 | 0.0004   |
| Analogy | 9594.363  | 9035.328 | 0.21   | 0.223    |
| Rotate  | 2327.429  | 1763.87  | 0.365  | 0.439    |
| SimplE  | 7837.303  | 7338.044 | 0.166  | 0.174    |

## 4.1 Results

Graph connectivity was effective, at least in our experiments. Indeed, the dataset's dispersion and heterogeneity, the graph's non-uniformity, the aggregation of the triples, and the degree of the node non-uniformity challenge any analysis and inference in the knowledge graphs.

While the only remarkable changes were removing relations of less than 100 and entities with less than five connections, the Raw MRR in the TransE model differed by about 1311. Other models' implementation results were similar to TransE. Two hypotheses could account for this. Hypothesis 1: The presence of entities rarely used in Wikipedia reduces the likelihood of selecting a new valid triple.

Some entities cause chain formation in the graph, making the knowledge graph and inferring and technically predicting from this graph a challenging task. Since the points they earn in the scoring function are low, they will not play a role in predicting the triples and will not be selected. These changes are shown in Table 4 and Figure 2.

Table 4: Farsbase and Farspredict triple frequency for entities

|                  | Sum of triples in a range | |
|------------------|----------|------------|
| Entity degree    | Farsbase | Farspredict|
| $n \leq 5$       | 754432   | 65834      |
| $5 < n \leq 50$  | 76233    | 2068       |
| $50 < n \leq 500$| 1273     | 183        |
| $500 \leq n$     | 13       | 100        |

Hypothesis 2: relations with a few triples are private ones used in a particular situation that cannot be generalized to other cases. Offering a triple with this type can reduce the link prediction accuracy. This threshold ignores relations with little effect on knowledge use and extraction, or instead, scattering and heterogeneity of the knowledge graph. These changes are shown in Table 5.

Overall, compared to standard dataset results, the results presented in Table 3 show that the mean rank of Farspredict link prediction is higher than standard datasets. This happened while our final dataset was approximately equal to Freebase. This difference in value in criterion "Mean Rank" could be because Farspredict is still sparse, and the number of entities is 107827, while the number of FB15K's entities is 15951. Although the number of triplets is not very different, the number of entities is about 7.2 times larger.

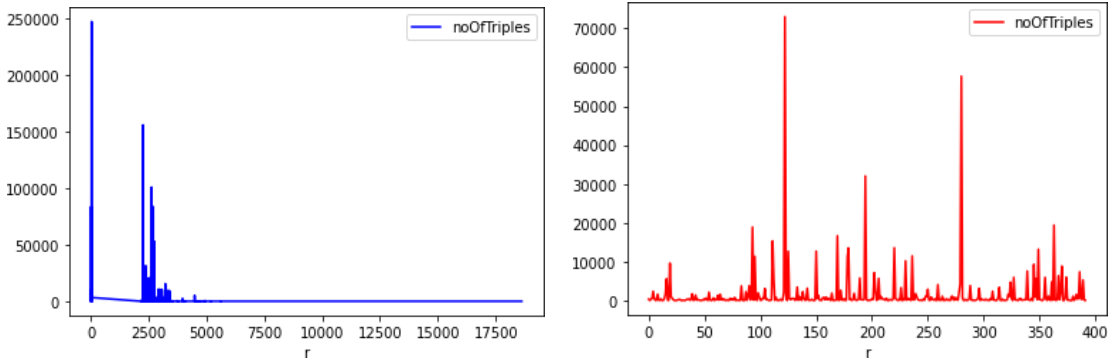

Figure 2: Triple numbers based on relation types. a) Farsbase, b) Farspredict

Table 5: Farsbase and Farspredict triples' frequency for relations

|  | Sum of triples in a range | |
| --- | --- | --- |
| #triples | Farsbase | Farspredict |
| $n \leq 100$ | 15810 | 141 |
| $100 < n \leq 1000$ | 436 | 174 |
| $1000 < n \leq 10000$ | 139 | 61 |
| $10000 \leq n$ | 45 | 16 |

## 5  Discussion

A knowledge graph completion is a valuable technique to ensure better knowledge extraction and inference. In this regard, link prediction through the KGE model is used for knowledge graph completion. Since the lack of using this tool in the Persian language can be seen, in this study, a knowledge graph was created for link prediction, and then its embedding was done in vector format with knowledge graph embedding models.

The results obtained in section 3 are equivalent to the outputs of the graph knowledge embedding models on Freebase, so the first hypothesis put forward for the similarity of implementing these models on Farspredict is well supported. Also, according to the results reported in Table 3, the deep learning method has worked well, and the second hypothesis has been proved.

In this project, a prospective investigation of factors and routines that transform our dataset into a link prediction in the knowledge graph was done. We found some steps for changing the graph structure that could elevate the quality of the proposed knowledge graph for link prediction. We perceived that graph connectivity as an essential factor statistically significantly reduces the mean rank of KGE models. Other factors that facilitated the mean rank factor were removing relations and entities with negligible dispersion. The positive association between decreasing the number of relation types and the mean rank factor is evident in this study. Our study includes the most significant prospective analysis of link prediction on the Persian knowledge date.

The embedded Persian knowledge graph and the suggested triples of the link prediction can be used in fields that work to extract knowledge, such as reasoning on knowledge graphs. Previous studies have reported many multilingual knowledge graphs and KGE models, which motivate us to do this project in Persian.

Despite the final analysis demonstrating the advantages of the new Persian knowledge graph, certain shortcomings may be tracked back to the sources used to extract the first triples. These sources are mainly from Persian Wikipedia pages whose contents have not been validated, and also similar verbs and words can be seen in many applications.

Consistent with previous studies, we observed that Farspredict triples' numbers are similar to Freebase, and the link prediction accuracy on embedded vectors is close to Freebase results and, in some cases, even better.

Researchers working on the knowledge graph completion might benefit from the proposed graph. We hope that by conducting additional studies and projects in this field, we achieve a more robust and complete Persian knowledge graph to be used as a knowledge reference in artificial intelligence projects.

# 6 Discussion

A Persian knowledge graph can be essential in some fields, including link prediction. Indeed, it was found that the only Persian knowledge graph (Farsbase) cannot meet the link prediction needs. We concluded that the main problem with graph structure is that it must be uniform, and the triples must be evenly distributed on the surface of the graph. In addition to the graph structure, removing several of Wikipedia's extracted triples whose components were in other languages, such as Arabic, and triples with two entities, were responsible for structural problems in the dataset, which ultimately weakened the link prediction results. Implementation of the KGE models on the last version of the knowledge graph reveals that using the KGE models for link prediction in other languages not only leads to excellent achievements but also pays attention to the quality of the knowledge graph, leading to better results than popular knowledge graphs.

Various changes in the knowledge graph dataset were evaluated by implementing the KGE models and using the mean rank and Top@10 factors. We hope our findings influence link prediction in the Persian knowledge graphs. Future work will entail refining the proposed KG by exploiting data from other sources.

By accessing the first Farspredict version, it is possible to do the following research and implementations in this area. Establishing a relationship hierarchy by getting ideas from Schema.org (as a reference taxonomy) can be a good candidate for future work. Moreover, combining Farspredict with valid knowledge graphs and creating multilingual knowledge graphs can be valuable work.

The knowledge graph can be exploited in any knowledge retrieval. However, it remains to be clarified whether our approach can be applied to knowledge graphs other than those employed in this study.

**Declarations**

- Funding: Not applicable
- Conflict of interest/Competing interests: The authors declare no conflict of interest.
- Ethics approval: Ethical approval is not applicable, because this article does not contain any studies with human or animal subjects.
- Consent to participate: Not applicable
- Consent for publication: Not applicable
- Availability of data and materials: The data included in this study are available upon request by contact with the first author.
- Code availability: Not applicable
- Authors' contributions: Conceptualization, N.T., and B.M.; methodology, N.T.; software, N.T.; validation, N.T.; formal analysis, M.J., and N.T.; investigation, N.T.; resources, N.T.; data curation, N.T.; writing—original draft preparation, N.T.; writing—review and editing, N.T., B.M., and M.J; visualization, N.T.; supervision, B.M.; All authors have read and agreed to the published version of the manuscript.